\documentclass[preprint,12pt, fleqn]{elsarticle}
\usepackage{amssymb,amsmath,array}
\usepackage{graphicx}
\usepackage[ruled,vlined]{algorithm2e}
\usepackage{algorithm2e,float}
\usepackage{hyperref}
\usepackage{subfig}
\usepackage[flushleft]{threeparttable}
\usepackage{tabularx}
\usepackage{tabularx,booktabs}
\usepackage{float}
\usepackage[english]{babel}
\usepackage{adjustbox}
\usepackage{blindtext}
\usepackage{tabularx}
 \usepackage{caption}
\usepackage[active]{srcltx}
\usepackage{array}
\usepackage{multirow}
\usepackage{booktabs}
\newcolumntype{L}[1]{>{\raggedright\let\newline\\\arraybackslash\hspace{0pt}}m{#1}}
\newcolumntype{C}[1]{>{\centering\let\newline\\\arraybackslash\hspace{0pt}}m{#1}}
\newcolumntype{R}[1]{>{\raggedleft\let\newline\\\arraybackslash\hspace{0pt}}m{#1}}

\usepackage{verbatim}
\usepackage{adjustbox}
\usepackage{amssymb}
\usepackage{lineno}

\journal{Applied Intelligence}

\begin{document}
	
	\begin{frontmatter}

	\title{Improved Multi-objective Data Stream Clustering with Time and Memory Optimization}
	
			\author[mymainaddress,mysecondaryaddress]{Mohammed Oualid Attaoui\corref{imoc-cor2}}
			\author[mymainaddress]{Hanene Azzag}
		\author[mysecondaryaddress]{Nabil Keskes}
		\author[mymainaddress]{Mustapha Lebbah}

        \cortext[cor2]{Corresponding author}

		\address[mymainaddress]{University Sorbonne Paris Nord, LIPN-UMR 7030 99 , av. J-B Cl\'ement, 93430 Villetaneuse, France\\\textbf{{attaoui, Mustapha.Lebbah, azzag}@lipn.univ-paris13.fr}}
		\address[mysecondaryaddress]{Higher School of Computer Science (ESI-SBA),  LabRI Laboratory, \\ Sidi Bel-Abbes, Algeria\\ \textbf{n.keskes@esi-sba.dz}}

		\begin{abstract}

		\indent  The analysis of data streams has received considerable attention over the past few decades due to sensors, social media, etc. It aims to recognize patterns in an unordered, infinite, and evolving stream of observations. Clustering this type of data requires some restrictions in time and memory. This paper introduces a new data stream clustering method (IMOC-Stream). This method, unlike the other clustering algorithms, uses two different objective functions to capture different aspects of the data. The goal of IMOC-Stream is to: 1) reduce computation time by using idle times to apply genetic operations and enhance the solution. 2) reduce memory allocation by introducing a new tree synopsis.  3) find arbitrarily shaped clusters by using a multi-objective framework.  We conducted an experimental study with high dimensional stream datasets and compared them to well-known stream clustering techniques. The experiments show the ability of our method to partition the data stream in arbitrarily shaped, compact, and well-separated clusters while optimizing the time and memory. Our method also outperformed most of the stream algorithms in terms of NMI and ARAND measures.
	\end{abstract}
	
	%
	%
	
		\end{frontmatter}
	


	\section{Introduction}
	
	\indent Swarm Intelligence or distributed intelligence is the collective behavior of decentralized and self-organizing natural or artificial systems \cite{nayyar2018advances}. It has attracted lots of interest from researchers in the last two decades due to their dynamic and flexible ability and that they are highly efficient in solving nonlinear problems in the real world \cite{nayyar2018introduction}. 

	\indent AntTree \cite{azzag2003anttree} is a hierarchical clustering method that models how ants form living structures and use this behavior to organize this data into a tree that is built in a distributed manner. Intuitively, each ant/data is located at the start of reliable support (tree root). The behavior of the ants then consists either in moving or in clinging to the structure to extend it and allow other ants to come and stick in their turn. This behavior is determined in particular by the similarity between the data and the local structure of the tree. The result is a tree-like organization of the data whose properties will allow us to determine a classification automatically and to have a visual overview of the tree.
	
	\indent The AntTree algorithm was proposed to deal with Data Stream, a kind of data that evolves and arrives in an unbounded stream. Analyzing data stream implies time and space constraints. The process of data stream clustering consists of creating compact and well-separated partitions from dynamic streaming data in only a single scan, using limited time and memory.
	
    \indent Most of the clustering techniques follow one objective function. However, every objective function represent a different property of the clusters, such as the compactness or the separateness of a cluster. When the algorithm assumes a homogeneous similarity measure over the entire data set, it becomes not robust to variations in cluster shape, size, dimensionality, and other characteristics \cite{handl2007evolutionary}. The Multi-Objective clustering methods (MOC) \cite{law2004multiobjective} retrieve clusters by applying two or more objective functions. It uses a two-step process: 1) Generate multiple clustering solutions and store the optimal ones. 2) Construct an optimal partition based on the Pareto-set solutions. The following definitions are useful to understand MOC methods :
	

	 \subsection{Definitions:}
	\begin{itemize}
		\item \textit{Dominated solutions:} a solution $X$ is said to dominate a solution $Y$  if $ \forall j = 1, 2, . . . , m, f_j (X) \leq f_j (Y)$, and there exists $k \in {1, 2,...,m}$ such that $f_k (X) < f_k (Y)$. Where $m$ is the number of objective functions and $f_j$ is the $j^{th}$ objective function. 
		\item \textit{Pareto-optimal solutions:} a solution $X$ is called Pareto-optimal if it is not dominated by any other feasible solutions. The set of non-dominated solutions is called Pareto-set.
		\item \textit{Idle times:} in the case of a slow stream, time delays between data points can appear e.g., times where no data point is available. Traditional algorithms will stop and wait for new data points to process them. Figure \ref{imoc-idle} illustrates the concept of idle times. 
		\begin{figure}[H]
			\centerline{\includegraphics[width=\textwidth]{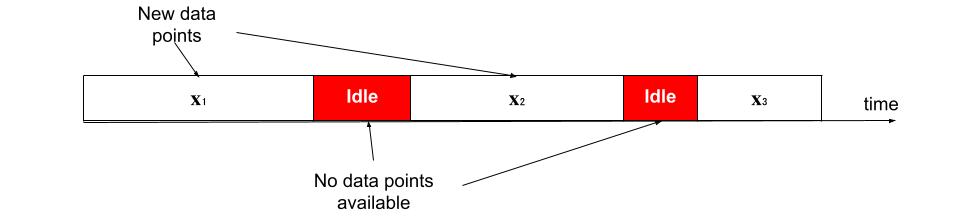}}
			\caption{\label{imoc-idle} Idle times.}
		\end{figure}
		
	\end{itemize}

	\section*{Main Contributions}
	\indent In our previous work \cite{attaoui2020moc} we proposed a multi-objective stream clustering method. This method uses genetic operators in every iteration to improve the solutions, costly in time computation. The second inconvenience of our previous method is calculating distances between all the clusters when trying to find the neighborhood of a particular cluster. This paper introduces an improved multi-objective clustering method based on data stream clustering, and Ant-Tree clustering algorithm \cite{azzag2003anttree}. This method optimizes the computation time by applying the genetic operators only in idle times to improve the solution instead of using them in every iteration. It optimizes memory allocation by using the Ant-Tree algorithm to find a cluster's neighborhood. It also introduces a new aggregation approach for the Ant-Tree algorithm to store only a synopsis of the data instead of storing all the data points. The method presented in this paper has the following merits compared to the other multi-objective and single-objective data stream clustering methods:
	\begin{itemize}
		\item It uses the Ant-tree algorithm to give the hierarchical aspect to our method and make it easier to determine the clusters' neighborhood. It also reduces memory allocation by modifying the AntTree algorithm not to store data points.
		\item It reduces computation time by using idle times to apply genetic functions and enhance clustering quality.  
		\item It optimizes two objective functions to obtain high-quality solutions and arbitrarily shaped clusters.
		\item It does not require the specification of the number of clusters and uses a Fading function to consider the most recent data as more important and reflect better the changes in the data distribution.
	\end{itemize}
	This paper is organized as follows: in section \ref{imoc-sec:related}, we present some background methods. In section \ref{imoc-sec:proposed}, we describe our method and its main features. In section \ref{imoc-sec:results}, we present the experiments and the results obtained and compare them to some known clustering methods. Finally, we conclude this paper.
	
	
	
	\section{Related Works}
	\label{imoc-sec:related}
	\indent This section discusses previous works on data stream clustering problems and highlights the most relevant algorithms proposed in the literature to deal with these problems. For stream clustering algorithms, only one Multi-Objective clustering method has been proposed. In \cite{paul2020online}, authors optimize multiple objectives capturing cluster compactness and feature relevancy. They consider an evolutionary-based technique and optimize multiple objective functions simultaneously to determine the optimal subspace clusters. The generated clusters in the proposed method are allowed to contain overlapping of objects. 
	
	The closest methods to MOC are Evolutionary algorithms since they use the same encoding of the solutions and the same process with a single objective function.
	\subsection{AntTree Clustering}
	\label{imoc-anttree}
	Ant-Tree algorithm \cite{azzag2003anttree} produces a hierarchical structure in an incremental manner like how the ants join together. In this algorithm, each ant represents a single data point, and it moves in the structure according to the similarity $Sim (i, j)$  with the other ants already connected in the tree under construction. $Sim (i, j)$ is represented by the euclidean distance between two ants $i$ and $j$. One should notice that this tree will not be strictly equivalent to a dendrogram as used in standard hierarchical
	clustering techniques: each node in our tree will correspond to one data while this is not the case in general for dendrograms, where data only correspond to leaves. \\
	
	Starting from the support, materialized by a fictitious node $f_0$, the ants will progressively fix themselves on this initial point, then successively on the ants set at this initial point, and so on until all the ants are attached to the structure.  During the construction of the structure, each ant $f_i$ is either moving on the graph or connected to it. In the first case, $f_i$ is free to move to a neighbor of the ant on which it is located (or to the support). 
	
	In the second case, $f_i$ will no longer be able to be released. Furthermore, we will consider the fact that each ant has only one outgoing link to other ants and cannot have more than $L_{max}$ links connected to it from other ants (tree having at most $L_{max}$ threads per node). Initially, all ants are placed on the $f_0$ support. They will each have a similarity threshold and a dissimilarity threshold, which are set to 1 and 0, respectively.\\
	
	An ant will connect under an existing node of the tree (ant $f_{pos}$ ) if it is sufficiently similar to this node but also dissimilar enough to the threads of the node: $f_i$ will thus form a subclass of $f_{pos}$ which will be different from the other subclasses of $f_{pos}$ (possibly already existing). Otherwise, $f_i$ will move randomly in the tree, looking for another location to fix itself. As the $f_i$ ant fails in its attempts to attach to the structure, it is made more tolerant in order to increase its chances of connecting to the next iteration concerning it: its similarity threshold is decreased, and its dissimilarity threshold is increased. The particular case of the $f_0$ support is treated as follows: an ant connects to the support if it is sufficiently dissimilar to other ants already connected directly to $f_0$. It means that a new class has just been built at the highest level of the tree. This class must be as distinct as possible from the other classes already created.\\
	
	The algorithm ends when all the ants are connected. The sub-trees appearing at the first level of the tree, just below the support, will be interpreted as different classes. The properties of the tree can be analyzed visually and interactively (e.g., the classification error decreases as one goes down the tree). It is also possible to transform this tree into a dendrogram (by scrolling down the data placed on internal nodes to leaves.  The Ant-Tree algorithm is presented in Algorithm \ref{anttree_algorithm} for a given ant $a_i$.
		\begin{sloppypar}	
\begin{algorithm}[H]
		\small
			
	           \If{No ant or only one ant is connected to the support $a_{pos}$}
	         {
	          - connect $a_i$ to $a_{pos}$ ;\
	          }
	          \ElseIf{Two ants are connected to the support}{
	           - Disconnect the second ant from $a_{pos}$ (and recursively all ants connected to it);\
               - Place all these ants back onto the support $a_0$;\
               - Connect $a_i$ to $a_{pos}$;\
	          }
	          \Else{
	          - let $T_{Dissim}(a_{pos})$ be the lowest dissimilarity value between daughters of $a_{pos}$ (i.e. $T_{Dissim}$($a_{pos}$) = $M$ in Sim($a_{j}$ , $a_{k}$) where $a_{j}$ and $a_{k}$ $\in$ {ants connected to $a_{pos}$});\
              - If $a_{i}$ is dissimilar enough to $a^{+}$ (Sim($a_{i}$ , $a^{+}$) < $T_{Dissim}$($a_{pos}$)) Then $a_{i}$ connects to $a_{pos}$;\
              - Else $a_{i}$ moves toward $a^{+}$;\
	          }

			\caption{Connection of an ant $a_i$ in the Ant-Tree}
			\label{anttree_algorithm}
		\end{algorithm}
	\end{sloppypar}

	\subsection{Evolutionary Multi-Objective Clustering Methods (MOC)}
In the past decade, multi-objective evolutionary algorithms have been heavily used in clustering because of their effectiveness. However, there has not been any dedicated effort to review all of these methods. The most prominent effort in this direction can be found in \cite{mukhopadhyay2015survey}, in which many multi-objective clustering algorithms and techniques were presented. This section presents a thorough survey of the state-of-the-art for a wide range
of multi-objective clustering algorithms.\\

\begin{figure}[H]
		\centerline{\includegraphics[width=\textwidth]{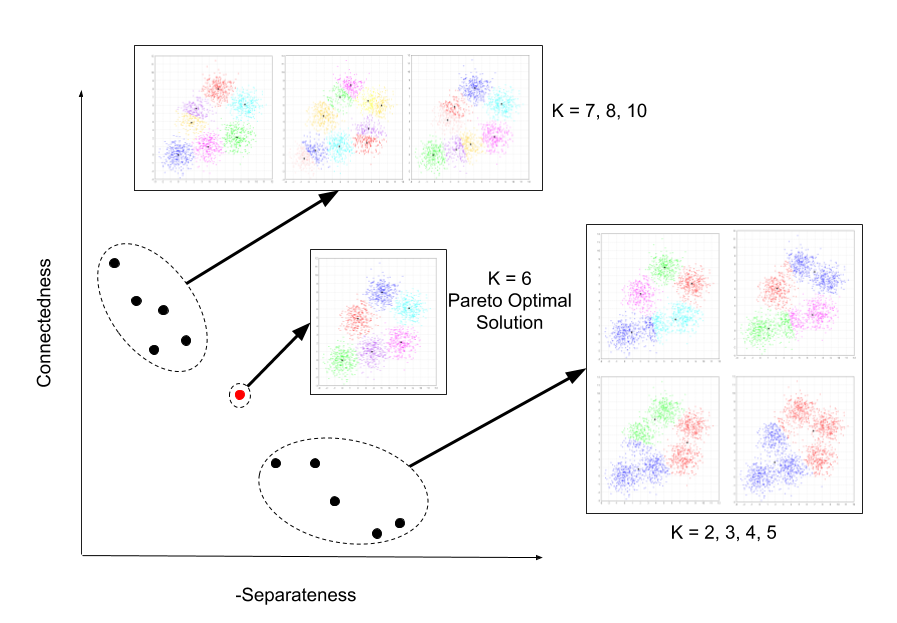}}
		\caption{\label{mock} Clustering solutions plotted according to their objective functions. Each point represents a clustering solution. The pareto optimal solution is obtained when K=6.}
\end{figure}

\indent This section discusses previous works on multi-objective clustering problems and highlights the most relevant algorithms proposed in the literature to deal with these problems.
\newline
\indent MOCK \citep{handl2007evolutionary} Multi-objective clustering with automatic K-determination, consists of two main phases: In its initial clustering phase, MOCK uses a Multi-Objective Evolutionary algorithm  (MOEA) to optimize two complementary clustering objectives. The output of this first phase is a set of a mutually non-dominated clustering solution. Each corresponds to different tradeoffs between the two objectives. In the second phase, MOCK analyzes the shape of the tradeoff curve. It compares it to the tradeoffs obtained for an appropriate null model (i.e., by clustering random data). Based on this analysis, the algorithm provides an estimate of the quality of all individual clustering solutions and determines a set of potentially promising clustering solutions. Often, a single solution is preferred, and, in these cases, the number of clusters inherent to the data set, $k$, is thus estimated implicitly. Figure \eqref{mock} illustrates the pareto set in MOCK algorithm. The improved version of MOCK, $\Delta$-MOCK has been proposed \citep{garza2017improved}, which can significantly decrease the computational overhand and reduce the search space. 

An Ant Colony Optimization-based clustering method ACO-C \citep{inkaya2015ant} combines the connectivity, proximity, density, and distance information with the exploration and exploitation capabilities of ACO in a multi-objective framework. The proposed clustering methodology is capable of handling several challenging issues of the clustering problem, including solution evaluation, extraction of local properties, scalability, and the clustering task itself.\\

Multi-objective evolutionary algorithms with simultaneous clustering and classification MOASCC \citep{luo2016learning} uses a clustering process to enhance the performance of the classification. To achieve this goal, two objective functions, fuzzy clustering connectedness function, and classification error rate, are adopted. A mutation operator is designed to make use of the feedback from both clustering and classification. \\

\begin{figure}[H]
		\centerline{\includegraphics[width=\textwidth]{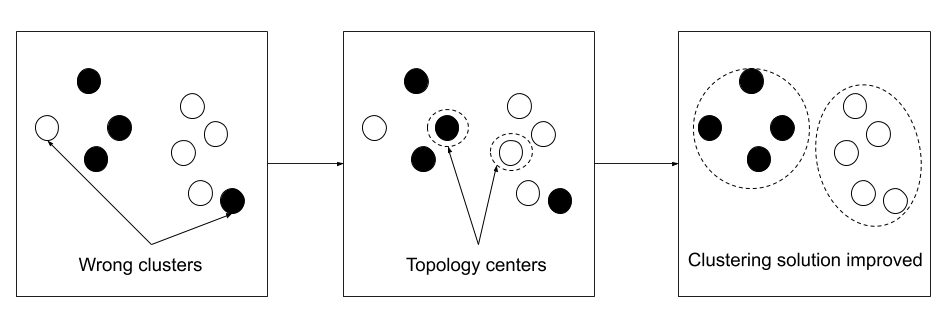}}
		\caption{\label{topological_center}Using topology centers to improve the clustering solution.}
\end{figure}

\indent IMCPSO \citep{gong2017improved} proposes an improved multi-objective clustering framework using particle swarm optimization. The authors used the overall deviation and mean distance between clusters as objective functions. They introduced a clustering method to improve each particle (clustering solution) by finding a topological center, which is the point that has the maximum neighbors belonging to the same cluster Figure \eqref{topological_center} illustrates the using of topological centers to improve the clustering. Finally, the best particle is selected from the Pareto-set based on the sparsity of the solution.  \\
EMO-KC \citep{wang2018multi} uses the term bi-objective clustering to describe a MOC method with two objective functions. The method has two main steps  (i) constructing two conflicting objective functions, and (ii) solving the bi-objective optimization problem with an effective EMO(Evolutionary Multi-Objective) algorithm.\\

Another MOC algorithm, SOMDEA-clust \citep{saini2019sophisticated}, proposes an efficient automated decomposition-based multi-objective clustering technique, which is a hybridization of Self-Organizing Maps (SOM) and differential evolution algorithm. Two internal cluster validity indices, namely, Silhouette index (SI) and PBM (Pakhira-Bandyopadhyay-Maulik) index, are used as objective functions. SOM algorithm is used to creates new solutions based on the neighborhood of each neuron.  AMOGA \citep{dutta2019automatic} Automatic clustering by a multi-objective genetic algorithm is a Multi-objective clustering algorithm that handles numeric and categorical features. Each clustering solution is encoded as a gene to apply Genetic operators. The method initializes a population using  K-prototypes algorithm and GA operators crossover and mutation. AMOGA uses compactness and separateness as objective functions and different validity measures (DB index, Purity...) to select the best solution from the Pareto-optimal set.

Multi-objective Gradient Evolution algorithm \citep{kuo2020multi}  extends the Gradient Evolution GE algorithm, so then it can be applied for the multi-objective problem. This paper applies the Pareto ranking assignment to sort the vectors based on their fitness values. K-means is then used to perform a final clustering on the Pareto-optimal solutions to obtain the final clustering.\\

Combinatorial Multi-Objective Pigeon Optimization algorithm (CMOPIO)  \citep{chen2020multi} is based on a bio-inspired algorithm called Pigeon Optimization PIO. In CMOPIO, pigeons only interact with the pigeons in their neighborhood. Meanwhile, the update of the pigeon's position and velocity relies on each pigeon's neighborhood rather than the global best position. These improvements allow the CMOPIO to identify a variety of Pareto optimal clustering solutions. \\
Table \eqref{moc-compar} Compares the Multi-Objective Clustering Algorithms.\\
\begin{table}[H]
\caption{Comparison Between Multi-Objective Clustering Algorithms}
\label{moc-compar}
\small

				\centering
\makebox[\textwidth]{\begin{tabular}{|c|c|c|c|}

\hline
Method                                                 
& Encoding scheme & Genetic functions                                                                        & Objective functions                                                                        \\ \hline
MOGE                                 & 
Centroid-based      & /                                                                                   & \begin{tabular}[c]{@{}c@{}}SSW(Separeteness)\\ SSB(Compactness)\end{tabular}               \\ \hline
CMOPIO                                                 
& Locus-based     & /                                                                                        & \begin{tabular}[c]{@{}c@{}}Connectivity\\ Compactness\end{tabular}                         \\ \hline
MOCK                                                  
& Locus-based     & /                                                                                        & Stability                                                                                  \\ \hline
$\Delta$-MOCK  
& Centroid-based  & \begin{tabular}[c]{@{}c@{}}Crossover\\ Mutation\end{tabular} & \begin{tabular}[c]{@{}c@{}}Connectivity\\ Class error rate\end{tabular}                    \\ \hline
ACO-C
& Point-based     & /                                                                                        & \begin{tabular}[c]{@{}c@{}}Adjusted Compactness\\ Relative Separateness\end{tabular}       \\ \hline
MCPSO  
& Locus-based     & /                                                                                        & \begin{tabular}[c]{@{}c@{}}Compactness\\ Separateness\end{tabular}                         \\ \hline
SOMDEA-Clust                                           
& Centroid-based  & \begin{tabular}[c]{@{}c@{}}Mutation\\ Crossover\end{tabular}                             & \begin{tabular}[c]{@{}c@{}}PBM Index\\ Silhouette Index\end{tabular}                       \\ \hline
IMCPSO    
& Locus-based     & /                                                                                        & \begin{tabular}[c]{@{}c@{}}Overall deviation\\ Mean distance between clusters\end{tabular} \\ \hline
MOEASCC 
& Centroid-based  & Mutation                                                                                 & \begin{tabular}[c]{@{}c@{}}$J_{In}$ (Connectedness)\\ $J_{Add}$ (Error rate)\end{tabular}  \\ \hline
EMO-KC
& Centroid-based  & \begin{tabular}[c]{@{}c@{}}Crossover\\ Mutation\end{tabular}                             & \begin{tabular}[c]{@{}c@{}}SSD\\ Overlap-Separateness\end{tabular}                         \\ \hline

\end{tabular}}

\end{table}

	\subsection{Data Stream Clustering Methods}
	\indent To the best of our knowledge, no Multi-Objective clustering method for data stream has been proposed. As discussed above, the closest methods to MOC are Evolutionary algorithms.  evoStream \cite{carnein2018evostream} (Evolutionary Stream Clustering) makes use of an evolutionary algorithm to bridge the gap between the online and offline components. Evolutionary algorithms are inspired by natural evolution where promising solutions are combined and slightly modified to create offsprings, which can yield an improved solution. By iteratively selecting the best solutions, an evolutionary pressure is created, which improves the result over time. \textbf{evoStream} \citep{carnein2018evostream} (Evolutionary Stream Clustering) makes use of an evolutionary algorithm to bridge the gap between the online and offline components. By iteratively selecting the best solutions, an evolutionary pressure is created, which improves the result over time. evoStream uses this concept to enhance the macro-clusters through recombinations and small variations iteratively. Since macro-clusters are created incrementally, the evolutionary steps can be performed while the online components wait for new observations, i.e., when the algorithm would usually idle. As a result, the computational overhead of the offline part is removed, and clusters are available at any time. The online component is similar to DBSTREAM \citep{hahsler2016clustering} but does not maintain a shared-density since it is not necessary for reclustering.
 
 \indent evoStream is based on \textbf{DBSTREAM} \citep{hahsler2016clustering} (Density-based Stream Clustering), which uses the shared density between two micro-clusters to decide whether micro-clusters belong to the same macro-cluster. A new observation is merged into micro-clusters if it falls within the radius from their center. Subsequently, the centers of all clusters that absorb the observation are updated by moving the center towards x. If the point is not assigned to a cluster, it is used to initialize a new micro-cluster. Additionally, the algorithm maintains the shared density between two micro-clusters as the density of points in the intersection of their radii, relative to the size of the intersection area. In regular intervals, it removes micro-clusters and shared densities whose weight decayed below a respective threshold. In the offline component, micro-clusters with high shared density are merged into the same cluster.
 
evoStream was used in \citep{supardi2020evolutionary} to detect outliers in a data stream. The goal of this method is to treat the distinct data object as an outlier detection problem rather than the categorization problem.

\indent \textbf{HDCStream} \citep{amini2014fast} (hybrid density-based clustering for data stream) first combined grid-based algorithms with the concept of distance-based algorithms. In particular, it maintains a grid where dense cells can become micro-clusters as known from distance-based algorithms (see Section 4). Each observation in the stream is assigned to its closest microcluster if it lies within a radius threshold. Otherwise, it is inserted into the grid instead. Once a grid-cell has accumulated sufficient density, its points are used to initialize a new micro-cluster. Finally, the cell is no longer maintained, as its information has been transferred to the micro-cluster. In regular intervals, all micro-clusters and cells are evaluated and removed if their density decayed below a respective threshold. Whenever a clustering request arrives, the microclusters are considered virtual points to apply DBSCAN \citep{ester1996density}. The algorithm consists of three steps: (1) Merging or mapping: the new data point is added to an existing mini-cluster or mapped to the grid. (2) Pruning Grids and Mini-clusters: the grids cells, as well as mini-cluster weights, are periodically checked in pruning time. The periods are defined based on the minimum time for a mini-cluster to be converted to an outlier. The mini-clusters with weights less than a threshold are discarded.  (3) Forming final clusters: final clusters are created based on mini-clusters, which are pruned. Each mini-cluster is clustered as a virtual point using a modified DBSCAN.

\textbf{FlockStream} is a bio-inspired algorithm for clustering data stream \cite{kennedy2006swarm} simulating the behavior of a group of birds in flight. Boid is the abbreviation of the word bird-oid (which means in the form of a bird). These boids are interacting and follow certain rules:

\begin{itemize}
\item \textbf{cohesion} to form a group, the boids are getting closer to each other
\item \textbf{separation} 2 boids can not be in the same place at the same time
\item \textbf{alignment} to stay grouped, boids try to follow the same path \\
\end{itemize}

FlockStream uses agents to mimic the behavior of boids. Each point is associated with an agent. An agent can be of three types: basic, p-representative (potential micro cluster), or co-representative (outlier microcluster, it can become p-representative if adding points, its weight exceeds a threshold). In the initialization phase, a set of basic agents is deployed in In space, the agents that have a great similarity approach (cohesion) form a cluster, while the other agents separate. The Euclidean distance is used to calculate the dissimilarity between agents. Agents can leave one group to join another with more similar agents. at the end of this phase, a summary for each cluster is calculated, and the other two types of agents appear p-representative and o-representative. In the second step, a mass of data stream is inserted. In this phase, the agents are updated as follows:

\begin{itemize}
\item if an o-representative or p-representative meets another representative, if their distance is less than a threshold, then they join to form a swarm (cluster)
\item a basic agent A meets a representative R, if their calculated distance is lower than a threshold, A is absorbed by R
\item a basic agent meets another, so if their similarity is less than a threshold, he joins to form an o-representative.
\end{itemize}
 We list the limitations and the merits of each algorithm in the following:
\begin{itemize}
\item \textbf{evoStream}:   \textbf{-Merits:} - Use idle times to improve the clustering quality - Output clusters at any time - Detection of outliers \textbf{-Limitations:} - Requires the set of the clusters number - Not suitable for high dimensional data.                   

\item \textbf{DBStream}:    \textbf{-Merits:} - Use the shared density between clusters to determine  if two clusters can be merged - Robust to noise \textbf{-Limitations:} - Several parameters need to be set - Depends on the insertion order of the data points.

\item \textbf{HDCStream}:   \textbf{-Merits:} - Handles outliers - Improves the computation time and quality                                            \textbf{-Limitations:} - Unable to detect variant levels of density - Can not handle high dimensional data.      
\item \textbf{FlockStream}: \textbf{-Merits:} - Detects outliers - lower time complexity \textbf{-Limitations:}  - Unable to handle high dimensional data.                         
\end{itemize}
	\section{Proposed Method}
	\label{imoc-sec:proposed}
	In this section, we introduce  IMOC-Stream (Multi-Objective AntTree Clustering data stream). The algorithm is based on AntTree clustering and combines stream clustering and multi-objective clustering to create a Multi-objective stream clustering algorithm that satisfies two objective functions. It makes use of the hierarchical nature of AntTree to improve the clustering quality. We describe in the following sections the main properties of IMOC-Stream.
 
	\subsection{Clustering in a Streaming Context}
	\indent We assume that the data stream consists in a sequence $\mathcal X = \{\mathbf x_1, \mathbf x_2, ..., \mathbf x_n\} $  of $n$ (potentially infinite) elements, arriving at times $t_1, t_2, ..., t_n$, where $\mathbf x_i = (x_{i}^1, x_{i}^2, ...,x_{i}^d )$. Since the most recent data points are more important and reflect better the changes in the data distribution, we use temporal windows to consider only recent data for the clustering. A set of $m$ clustering solutions $\mathcal S$ is generated and updated for each window $\mathcal S = {\mathcal C_1, \mathcal C_2,..., \mathcal C_m}$ where $\mathcal C_j$ is the $j^{st}$ clustering solution and is represented by K clusters $\mathcal C_j = c_1,c_2,..,c_K$. Each cluster $c$ is represented by a prototype $\mathbf w_c$ where $\mathbf {w}_c= (w_c^1, w_c^2, \dots, w_c^d)$ and $d$ is the dimension of the data. Each cluster is associated with a weight $\pi_{c}$ that decreases over time based following a fading function. 

	When the first batch of data arrives in the first time window, we create the tree as a clustering solution according to Section \ref{imoc-anttree}, and this solution is stored in the Pareto-set. From the same batch of data we initialize several solutions using K-means \cite{pelleg2000x} with different $K$, GNG \cite{fritzke1995growing}, DBScan \cite{ester1996density}. The parameter settings of these algorithms are reported in Table \ref{imoc-settings}. The generated solutions are combined with the tree solution by the mutation and crossover operators, and the results are added to the solutions-set. We compute the objective function values for each solution-set and store the non-dominated solutions in the Pareto-set. 
	\begin{table}[H]
\centering
\begin{tabular}{|c|c|c|}
\hline
\textbf{Algorithm} & \textbf{Parameters}                                                         & \textbf{Source}                                                                        \\ \hline
GNG       & $epochs$ = 30                                                         & Smile Package\footnote{https://haifengl.github.io}          \\ \hline
DBSCAN    & \begin{tabular}[c]{@{}c@{}}$minPts$ = 20\\ $radius$ = 10\end{tabular} & Smile Package\footnote{https://haifengl.github.io}          \\ \hline
K-means   & K vary from 2 to 15                                                   & Clustering4Ever\footnote{https://github.com/Clustering4Ever} \\ \hline
Ant-tree  &    $L_{max} = 10$                                                                   & Clustering4Ever\footnote{https://github.com/Clustering4Ever} \\ \hline
\end{tabular}
\caption{Parameter settings for the used algorithm}\label{imoc-settings}
\end{table}
	After the Initialization phase and for each new window of data points, each point in the current window is assigned to the closest center $c_{ij}$ in each clustering solution $C_j$ in the pareto-set. The  distance calculated between the data points and the centers is the euclidean distance. We note that for each clustering solution, a point can be assigned to only one cluster. After all points being assigned, we update the clustering solutions with the new assigned points. We compute the objective functions values $f_i$ and we update the pareto-set. If the system idles, the method combines the solutions in the pareto-set using the genetic operators and calculates the objective values of the new generated solutions. At the end of each iteration, the pareto-set contains a set of non-dominated solutions. At the end of the process, a set of non-dominated clustering solutions is stored. These solutions are equally good mathematically. We used an internal quality measure Davies Bouldin \cite{davies1979cluster} to select the best solution among the Pareto-set.\\
	
	
	\subsubsection{AntTree with Tree Aggregation}
	To deal with the memory constraints encountered when analyzing data streams, we propose a new representation of the tree to prevent storing all the data points and to reduce the memory allocation. The tree is initialized from the data points in the first window following the AntTree algorithm described in section \ref{imoc-anttree}. After placing all the points, we compute the prototypes $w$ of each cluster as the average of the points assigned to this cluster. All the points are discarded, and only the tree with the prototypes is stored in the memory. For the next windows, we assign the new data points to each cluster and update the prototype $\mathbf w_{c}$ as follows:
	\begin{equation}
	\mathbf w_{c}^{ (t+1)}=\frac{\mathbf w_{c}^{ (t)}n_{c}^{ (t)}\gamma + \mathbf z_{c}^{ (t)}m_{c}^{ (t)}}{n_{c}^{ (t)}\gamma + m_{c}^{ (t)}} 
	\label{imoc-updateproto}
	\end{equation} 
	Where $\mathbf w_{c}^{ (t)}$  is the previous prototype, $n_{c}^{ (t)}$ is the number of points assigned to the cluster, $\mathbf z_{c}^{ (t)}$  is the new prototype computed only from the current window.  $m_{c}^{ (t)}$ is the number of points assigned to the cluster $c$ in the current window:   $n_{c}^{ (t+1)} = n_{c}^{ (t)}+m_{c}^{ (t)} $. $\gamma$ is the decay factor that decreases over time to give more importance to most recent data $0<\gamma<1$. If $\gamma$ = 1 all data will be used from the beginning; $\gamma$ = 0 only the most recent data will be used.\\
	
\begin{figure}[H]
		\centerline{\includegraphics[width=\textwidth]{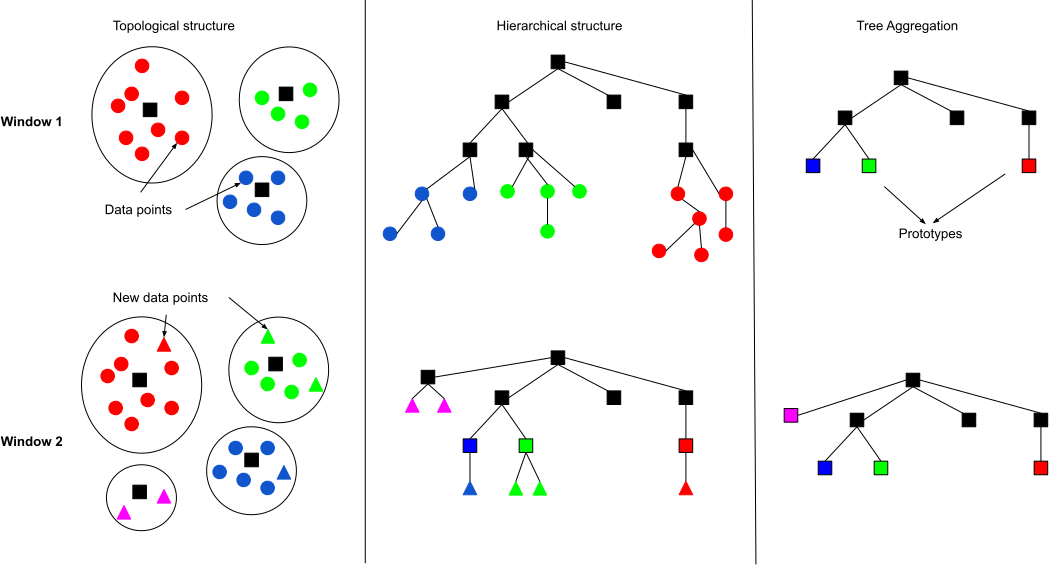}}
		\caption{\label{imoc-representation} Topological and Hierarchical representation and tree aggregation process. The circles represent the data points, the squares represent the prototypes and the triangles represent the new data points from the current window.}
	\end{figure}

	If a point is not assigned to a cluster, it becomes a prototype of a newly created cluster. Figure \ref{imoc-representation} illustrates tree representation and aggregation.    
	\subsubsection{Fading Function }
	Most data stream algorithms consider the most recent data as more important and reflect better the changes in the data distribution. For that, we consider a $Fading$ function, in which the weight of each cluster decreases exponentially with time $t$ by introducing a decay factor parameter $0<\gamma<1$.  
	\begin{equation}
	\pi_{c}^{ (t+1)}=\gamma \pi_{c}^(t),
	\label{imoc-nodeweightupdate}
	\end{equation} 
	where $n_c$ is the number of points assigned to the cluster $c$ at the current time $t$. If the weight of a node is below a threshold value, this cluster is considered outdated and removed.

	\subsection{Evolutionary Representation and Functions}
	Most of the Multi-Objective clustering methods use an evolutionary representation for the clustering solutions as their use of population enables the variation of solutions and makes it easier to keep a population of clustering solutions and apply genetic operators. However, the use of such representation requires the following concepts:
	\begin{itemize}
		\item Choosing an evolutionary encoding to represent a clustering solution.
		\item The generation of the initial population by an effective initialization scheme.
		\item Suitable genetic operators to mix the solutions.
		\item Choosing two or more objective functions as a fitness function to choose the non-dominated solutions.
		\item Developing a technique to obtaining a single clustering solution for the Pareto-set (leader selection method).
	\end{itemize}
	The choice of these components is crucial for the clustering quality and the algorithm scalability. In the next sections, we describe the components we chose after extensive experiments to deal with the requirements presented above.
	
	\subsubsection{Genetic Representation}
\begin{figure}[H]
		\centerline{\includegraphics[width=\textwidth]{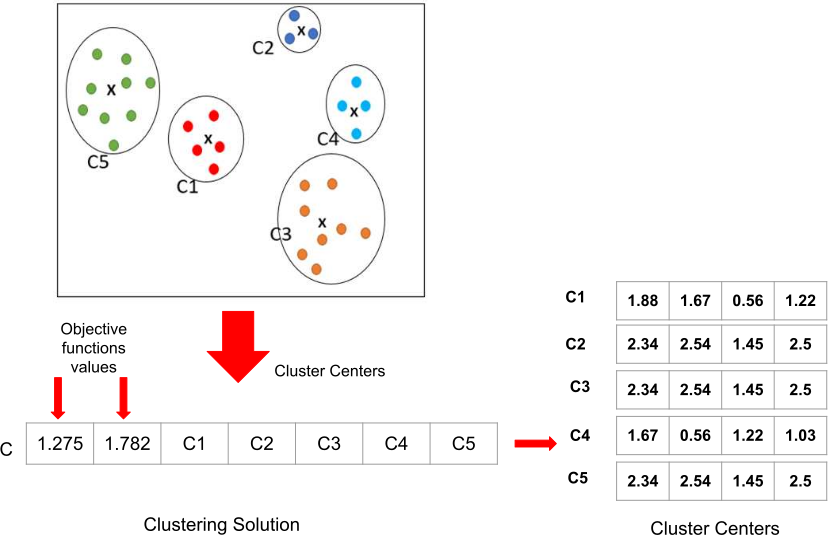}}
		\caption{\label{imoc-encoding} Clustering solution representation and conversion.}
	\end{figure}

	\indent Many representations were presented in the previous MOC methods \cite{mukhopadhyay2015survey}. However, these representations are not suitable for data stream clustering since data points can not be stored and have to be processed in one pass. Therefore, we chose a new genetic representation that facilitates the clustering update as the data flows. Each clustering solution is represented by a chromosome, which is an array of $K \times d$ + 2, where $d$ is the dimension of the data. The first and second components are the objective values for this solution. The last $K$ components represent the clusters. Each cluster is represented by a prototype $\mathbf{w}$ of $d$ elements. Figure \ref{imoc-encoding} illustrates the clustering representation and conversion.
	\subsubsection{Population Initialization}
\begin{figure}[ht!]
		\centerline{\includegraphics[width=\textwidth]{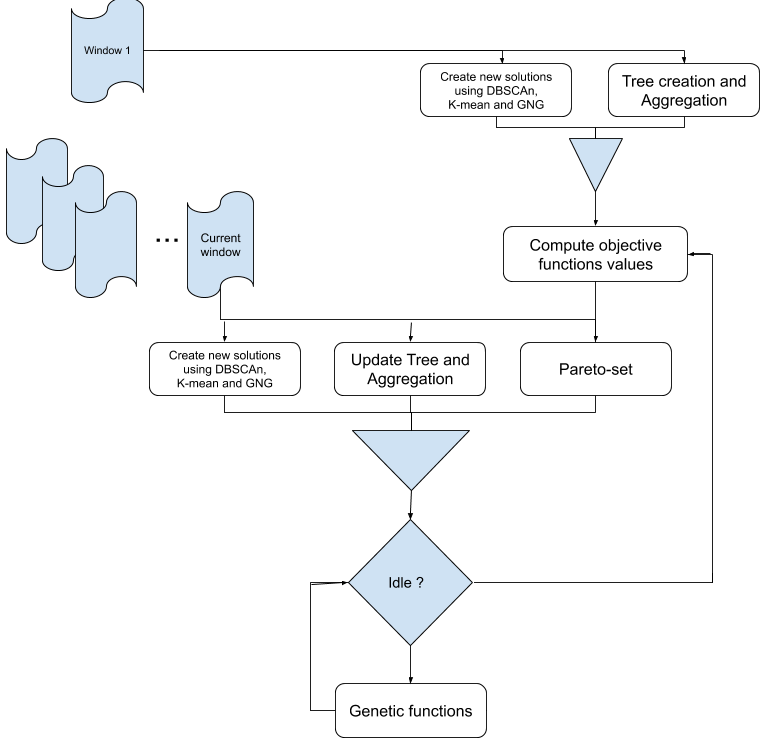}}
		\caption{\label{imoc-initialization} Initialization and update scheme.}
		
	\end{figure}

	\indent In each time window of the data stream, a set of clustering solutions is created and stored. Our algorithm does not require these solutions to have the same number of clusters. A first population is created from the first window using the AntTree algorithm combined with other solutions generated by several algorithms (K-means \cite{pelleg2000x} with different $K$, GNG \cite{fritzke1995growing}, DBScan \cite{ester1996density}). Those algorithms were chosen after extensive experimentation due to their ability to do a local search. The solutions given are encoded following the scheme described in Figure \ref{imoc-encoding}. We select the best solutions from this population to create new clustering solutions following the genetic operators Crossover and Mutation described in section \ref{imoc-operators}. After the first population initialized, we compute objective functions for each clustering solution and store the Pareto-optimal solutions into the Pareto-set. For each window of the data stream, the new data points belonging to the current window are used to update the solutions in the Pareto-set and to create new solutions. The Pareto-set is then updated with the non-dominated solutions. We describe the initialization and update scheme in Figure \ref{imoc-initialization}.
	\subsubsection{Genetic Functions}
	\label{imoc-genetic-ops}
	\label{imoc-operators}
	\indent Genetic operators are essential for MOC methods as they enable the variety and diversity of the clustering solutions. For our method, we use two genetic operators: Crossover and Mutation, to explore more solutions. The use of those operators helps find a better solution by combining the optimal solutions obtained from the other algorithms. 
	
	\begin{itemize}
		
		\item \textit{Crossover}:
		We used the single point crossover \cite{whitley1994genetic} in this paper due to its Independence of the ordering of genes. The goal of the crossover operator is to create new clustering solutions from the two-parent solutions. First, we randomly select the Pareto-set two solutions that have respectively $K_1$ and $K_2$ clusters. We choose randomly a crossover point $i$, as the number of clusters may vary, $i$ must satisfy $1<i<min(K_1,K_2)$.

        The first resulted clustering solution from the crossover is composed of cluster centers from $1$ to $i$ of the solutions with $min(K_1, K_2)$ cluster centers, and $i$+$1$ to $K$ of the second clustering solution. The second resulted clustering solution is composed of the cluster centers $i+1$ to $min(K_1,K_2)$ from the first solution and of cluster centers $1$ to $i$ from the second solution. Figure \ref{imoc-crossover} explains the process of crossover of two clustering solutions. 
		\begin{figure}[ht!]
			\centerline{\includegraphics[width=0.8\textwidth]{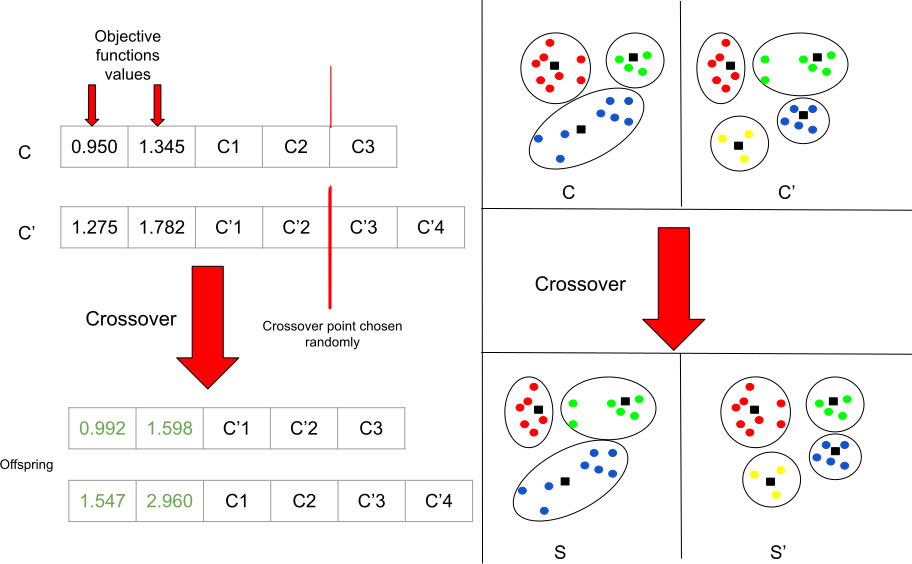}}
			\caption{\label{imoc-crossover} Crossover of two clustering solutions. The figure on the left represents the prototypes and the one on the right represents the topological clusters, the squares represent the prototypes and the circles are the data points. The data points are added to illustrate, in the clustering process, no data point is kept in the memory.}
			
		\end{figure}
		\item \textit{Mutation}:
		We use the random resetting mutation operator \cite{mitchell1998introduction} to change randomly some values of a cluster center of a clustering solution to explore global solutions. We select a clustering solution $C$ from the Pareto-set, then from each cluster center in $C$, we randomly select $\mu$ position values. For a value $v$, a number $0<\varrho<1$ is generated and the value $v$ is updated as follows:
		\begin{equation*}
		    v \pm \varrho * v,
		\end{equation*}
		 The '+' or '-' signs occur with equal probability. Figure \ref{imoc-mutation} illustrates the process of mutation of a clustering solution.
		\begin{figure}[H]
			\centerline{\includegraphics[width=0.8\textwidth]{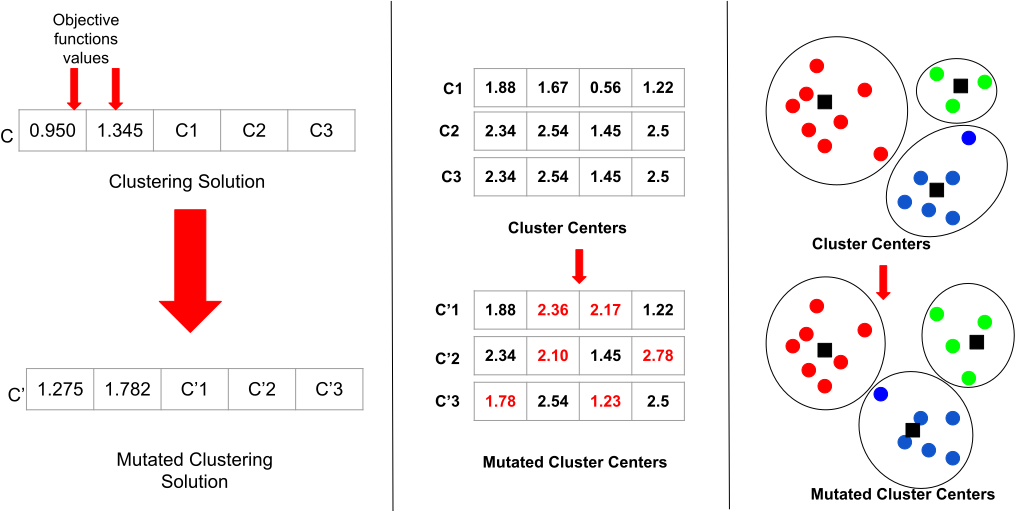}}
			\caption{\label{imoc-mutation} Mutation of a clustering solution. $\mu$ = 50\%}
			
		\end{figure}
	\end{itemize}
	Both operators are applied during idle times on the solutions from the Pareto-set. The solutions are selected based on their fitness score, equal to ($-separateness + compactness$). We select $\sigma$ clustering solutions and apply the genetic operators.
	\subsection{Objective Functions}
	One of the important aspects of MOC is the choice of suitable objective functions that are to be optimized simultaneously. For each clustering solution, several quality measures exist. The the goal is to have distinct clusters (separateness) that are the most dense in terms of the data points they contain (compactness). To satisfy these requirements, we introduce two objective functions $compactness$ and $separateness$. The combination of both objective functions allows us to have arbitrarily shaped clusters.
	\begin{itemize}
		\item \textit{Compactness}:
		the compactness of a clustering solution reflects the overall intra-cluster size of the data and has to be minimized. The compactness of a clustering solution in a streaming context is computed as follows:
		\begin{equation}
		Compactness_{\mathcal  C}^{t+1} = \gamma Compactness_{\mathcal  C}^{t} + \sum_{\mathbf x_i \in \mathcal X^{ (t+1)}}   \delta (\mathbf x_i,\mathbf w_{\phi (\mathbf x_i)})
		\label{imoc-compactness}
		\end{equation}
		Where $\mathcal X^{ (t+1)}$ is the current window and $\phi (\mathbf x_i)$ is the index of the cluster where $\mathbf x_i$ belongs. $\delta (\mathbf x, \mathbf w_{\phi (\mathbf x_i)})$ is the euclidean distance between the data point $\mathbf x$ and $\mathbf w_{\phi (\mathbf x_i)}$. $\gamma$ is the decay factor that decreases over time to give more importance to most recent data. The points of the previous windows are not kept, $Compactness_{\mathcal  C}^{t}$ has been computed in the previous window with the previous prototype $\mathbf w$. 
		
		
		\item \textit{Separateness}:
		the separateness of a clustering solution is the mean distance between clusters.  It reflects the inter-cluster similarity and should be maximized. The separateness of a cluster is the shortest distance between a data point in this cluster and another data point of his neighborhood belonging to another cluster. In a streaming context, the separateness is computed as follows:
		\begin{equation}
		Separateness_{\mathcal  C} = \frac{1}{|\mathcal  C|} \sum_{c \in \mathcal C} (min_{\mathbf x_i \in  c,   \mathbf x_j \in \mathcal K_{i},\mathbf x_{j} \notin c } \delta (\mathbf x_i, \mathbf x_j))
		\label{imoc-separateness}
		\end{equation}
		Where $\mathcal{K}_i$ is the neighborhood of the data point $\mathbf x_i$ belonging to the cluster $c$. The neighborhood of a node is determined through the AntTree method. The neighborhood of a cluster is the directly connected nodes to this one on the tree. 
	\end{itemize}

	\subsection{Solution Selection}
	\label{imoc-selection}
	At the end of the online phase, a set of non-dominated solutions is stored in the Pareto-set. These non-dominated solutions are equally good mathematically. We used an internal quality measure Davies Bouldin \cite{davies1979cluster} to select the best solution among the Pareto-set. The choice of an internal index is because the data might not be labeled. We sort all the solutions by their fitness (internal measures values), and we choose the best one as an output of the algorithm. The Davies-Bouldin index helps identify sets of clusters that are compact and well separated. The Davies-Bouldin index is calculated as:
\begin{equation}
    DBI = \frac{1}{K}\sum_{i=1}^{K} max_{i, j=1,..,K; j \neq i}{\frac{d(x_i, C_i) + d(x_j, C_j)}{d(C_i, C_j)}} 
\end{equation}
$d(x_i, C_i)$ is the distance between the data point $x_i$, and its cluster $C_i$ and K is the number of clusters. DBI varies between 0 (best clustering) and $+\infty$ (worst clustering).
	
	\subsection{Improved MOC-Stream Algorithm}
	IMOC-Stream is an extension of Multi-objective clustering for data stream to optimize computation time and memory allocation. It starts with creating a first clustering solution using the AntTree algorithm. In the contrast to the original algorithm where all the data points are stored, we introduced a new tree aggregation method to store only a synopsis of the data. The clustering solution is encoded and combined with different solutions obtained by different algorithms to create a population of solutions. The objective function values are computed for each solution, and only the non-dominated solutions are added to the Pareto-set. Then, we apply crossover and mutation on the best solutions selected from the Pareto-set and add the obtained solutions to the population. For each time window, the next point from the stream is mapped into the tree, the prototypes are computed, and only the aggregated tree is stored. We update the weights of the nodes and remove the outdated ones. If the stream idles, we apply genetic operators to generate more solutions. At the end of each time window, we compute the objective function values, select the non-dominated solutions, and update the Pareto-set. In the offline phase, we compute the Davies Bouldin index values of each potential solution and select the optimal one as an output for this algorithm. In summary, the algorithm of IMOC-Stream presented in this paper is described in Algorithm \ref{imoc-mocstream}.\\
	\begin{algorithm}[H]
		\SetAlgoLined
		\KwResult{Optimal clustering solution}
		From the first window: initialize the tree using AntTree algorithm and perform tree aggregation following Figure \ref{imoc-representation}\;
		Generate several clustering solutions using K-means with different K, GNG, and DBScan with different parameters\;
		Encoding the clustering solutions following scheme in Figure \ref{imoc-encoding}\;
		Apply Crossover and Mutation  following Figures \ref{imoc-crossover} and \ref{imoc-mutation} respectively. Add new solutions to the population of chromosomes\;
		Compute objective functions following equations \eqref{imoc-compactness} and \eqref{imoc-separateness}. Store non-dominated solutions in the Pareto-set\;
		
		\While{There is data available}{
			Map each point into the tree and compute prototypes following Equation\eqref{imoc-updateproto}\;
			For each clustering solution in the pareto-set, assign each point to the closest cluster\;
			Update each cluster in each clustering solution using the new points assigned as described in Equation \eqref{imoc-updateproto}\;
			Update weights of nodes following Equation \eqref{imoc-nodeweightupdate} and remove the outdated nodes\;
			Compute objective functions of the clustering solutions in the pareto-set and the new solutions generated. Update the pareto-set with the new non-dominated solutions \;
			\While{Idle}{
				Select best clustering solutions from Pareto-set based on their objective values \;
				Apply Crossover and Mutation following Figures \ref{imoc-crossover} and \ref{imoc-mutation} respectively. Add new solutions to the population of clustering solutions\;
			}
		}
		Select best solution among the pareto-set solutions as described in section \ref{imoc-selection}
		\caption{Improved MOC-Stream Algorithm\label{imoc-mocstream}}
	\end{algorithm}
	\section{Experiments and Discussion}
	\label{imoc-sec:results}
	\subsection{Datasets and Quality Criteria}
	The IMOC-Stream method described in this article was implemented in Scala programming language and will be available on Clustering4Ever GitHub repository\footnote{https://github.com/Clustering4Ever/Clustering4Ever}. We evaluated the clustering quality of  IMOC-Stream on several real \cite{uci} and synthetic\footnote{https://www.sites.google.com/site/nonstationaryarchive/} datasets.  We describe the datasets in Table \ref{imoc-datasets}. The mutation rate $\mu$ is set to 20\% and the number of selected clustering solutions to the crossover and mutation is set to $10$. 
	
	\begin{table}[htbp!]
		\centering
		\begin{tabular}{|c|c|c|c|c|}
			\hline
			\textbf{Dataset}     & \textbf{Instances} & \textbf{Features} & \textbf{Classes} & \textbf{$|Window|$} \\ \hline
			powersupply & 29,928    & 2        & 24 &  100    \\ \hline
			HyperPlan   & 100,0000  & 10       & 5  &    1000 \\ \hline
			Covertype   & 581,102   & 10       & 23  &   1000 \\ \hline
			Sensor      & 2,219,802 & 4        & 54   &   10000 \\ \hline
			1CDT      & 16000 & 2        & 2   &   100 \\ \hline
			1CSurr      & 55280 & 2        & 2   &   1000 \\ \hline
			4CR      & 144000 & 2        & 1   &   1000 \\ \hline
			GEARS-2C-2D      & 200000 & 2        & 2   &   10000 \\ \hline
		\end{tabular}%
		\caption{Description of datasets used in experimentation}\label{imoc-datasets}
	\end{table}
	
	For the quality measures, we used the internal measures (NMI) \cite{strehl2002cluster} and the Adjusted Rand index (ARAND) \cite{hubert1985comparing}, these two measures require the ground truth of the data to be available. $NMI$ provides a measure that is independent of the number of clusters as compared to purity. It reaches its maximum value of 1 only when the two sets of labels have a perfect one-to-one correspondence. The NMI of a clustering solution $C$ is calculated as follows :
	
	\begin{equation}
	NMI (Y, C) = \frac{2 \times I (Y; C)}{H (Y) + H (C)}
	\end{equation}
	Where $Y$ are true labels and $C$ are labels predicted by the algorithm. $I (Y; C) = H (Y) - H (Y|C)$ and $H (C)$ is the entropy of the partition calculated as follows: 
	\begin{equation}
	\sum_{k=1}^{K} \frac{n_k}{N}log (\frac{n_k}{N})
	\end{equation}
	Where $n_k$ is the number of points assigned to the partition $k$.
	ARAND index is a measure of agreement between two partitions: one given by the clustering process and the other defined by external criteria. Given the following contingency matrix, where $n_{ks}$ represents the number of points assigned to both clusters $k$ and $l$ of partitions $C$ and $C^{'}$:
	\begin{table}[htbp!]
		\small
		\centering
		\begin{tabular}{|c|c|c|c|c|c|}
			\hline
			& $Y_1$ & $Y_2$ & ... & $Y_s$ & Sums \\ \hline
			$X_1$ & $n_{11}$ & $n_{12}$ & ... & $n_{1s}$ & $a_1$ \\ \hline
			$X_2$ & $n_{21}$ & $n_{22}$ & ... & $n_{2s}$ & $a_2$ \\ \hline
			... & ... & ... & ... & ... & ... \\ \hline
			$X_r$ & $n_{r1}$ & $n_{r2}$ & ... & $n_{rs}$ & $a_r$ \\ \hline
			Sums & $b_1$ & $b_2$ & ... & $b_s$ &  \\ \hline
		\end{tabular}%
		\caption{Contingency matrix between two partitions $C$ and $C^{'}$ of $r$ and $s$ clusters respectively.}\label{imoc-contingency}
	\end{table}
	The Adjusted RAND index is calculated as follows:
	\begin{equation}
	ARAND = \frac{ \sum_{ij} { {n_{ij}}\choose{2} } - [ \sum_{i} { {a_{i}}\choose{2} } \sum_{j} { {b_{j}}\choose{2} } ] / { {n}\choose{2} } } { \frac{1}{2} [ \sum_{i} { a_{i}\choose{2} } + \sum_{j} { {b_{j}}\choose{2} } ] - [ \sum_{i} { {a_{i}}\choose{2} } \sum_{j} { {b_{j}}\choose{2} } ] / { {n}\choose{2} } } 
	\end{equation}

	\subsection{Experimental Settings}
	Assuming large high-dimensional data arrives as a continuous stream,  IMOC-Stream divides the streaming data into batches and processes each batch continuously. The batch size depends on the available memory and the size of the original dataset the size of the window for each dataset is shown in Table \ref{imoc-datasets}. We set the time interval between two batches to 1 second and the parameter $\gamma$ to 0.7.  \\ To show the effectiveness of our method, we compared it to five well known stream algorithms: $Stream KM++$ \cite{de2011extending}, $DStream$ \cite{tu2009stream}, $DBStream$ \cite{hahsler2016clustering}, $DenStream$  \cite{cao2006density} and $CluStream$ \cite{aggarwal2003framework} from R package streamMOA\footnote{https://github.com/mhahsler/streamMOA}. We repeated our experiments with different initialization and have chosen those giving the best results. Table \ref{imoc-parameters} shows the optimal parameter configurations.
	\begin{table}[htbp!]
		\centering
		\begin{tabular}{|c|c|c|}
			\hline
			\textbf{Algorithms} & \textbf{Parameters} & \textbf{Initialization} \\ \hline
			DStream & \begin{tabular}[c]{@{}c@{}}$gridsize$\\ $\lambda$\\ $gaptime$\\ $Cm$\end{tabular} & \begin{tabular}[c]{@{}c@{}}0.9\\ 0.001\\ 1000\\ 3\end{tabular} \\ \hline
			DBStream & \begin{tabular}[c]{@{}c@{}}$r$\\ $\lambda$\\ $gaptime$\\ $Cm$\end{tabular} & \begin{tabular}[c]{@{}c@{}}1.8\\ 0.001\\ 1000\\ 2.5\end{tabular} \\ \hline
			DenStream & \begin{tabular}[c]{@{}c@{}}$\epsilon$\\ $\mu$\\ $\beta$\end{tabular} & \begin{tabular}[c]{@{}c@{}}0.4\\ 1.605\\ 0.275\end{tabular} \\ \hline
			CluStream & $t$ & 2 \\ \hline
		\end{tabular}%
		
		\caption{Optimal parameter configurations for the algorithms used for the experimentation. For IMOC-Stream, the decay factor is fixed to 0.7.}\label{imoc-parameters}
		
	\end{table}

	\subsection{Clustering Evaluation}
	The results of IMOC-Stream on the datasets described above compared to the different algorithms are reported in Tables \ref{imoc-results} and \ref{imoc-results2}. The value of NMI and ARAND is the average value of ten runs. It is noticeable that IMOC-Stream gives better results than all the other methods. These results are due to the fact that our method optimizes two objective functions to maximize intra-cluster similarity and minimize inter-cluster similarity at the same time, which gives us a compact and well-separated clusters. The use of different algorithms to create a population of solutions allow IMOC-Stream to explore better solutions and escape the local minima.
	Another critical point is the use of the genetic parameters to combine the best solutions and explore the potential local solutions. The other algorithms are sensitive to the initialization of the settings, which justify why our algorithm yields better results since it has no input parameters. Finally, we noticed that the DStream algorithm gives better results compared to the other algorithms used in this experimentation since it is adapted to large datasets.\\
	
	 For synthetic datasets, IMOC-Stream also gave better results than the different stream algorithms except for StreamKM++ on the $1CSurr$ dataset. These results are due to the optimal choice of the $K$ for StreamKM++, which makes it find the exact number of clusters with synthetic datasets and gives better results. The number of clusters is not predefined in IMOC-Stream, but it still manages to find approximately the right amount of clusters. 
	\begin{table}[H]
		\begin{center}
			
			\resizebox{\textwidth}{!}{\begin{tabular}{|c|c|c|c|c|c|c|c|}
					\hline
					Dataset & Metrics & IMOC-Stream & StreamKM++ & DStream & DBStream & DenStream & CluStream \\ \hline
					powersupply & \begin{tabular}[c]{@{}c@{}}NMI\\ ARAND\end{tabular} & \begin{tabular}[c]{@{}c@{}}\textbf{0.466 $\pm$ 0.03}\\ \textbf{0.144 $\pm$ 0.03}\end{tabular} & \begin{tabular}[c]{@{}c@{}}0,232 $\pm$  0,05\\ 0,034 $\pm$ 0,01\end{tabular} & \begin{tabular}[c]{@{}c@{}}0,403 $\pm$  0,06\\ 0,049 $\pm$  0,01\end{tabular} & \begin{tabular}[c]{@{}c@{}}0,056 $\pm$ 0,01\\ 0.001 $\pm$ 0.00\end{tabular} & \begin{tabular}[c]{@{}c@{}}0.055 $\pm$ 0.01\\ 0.002 $\pm$ 0.00\end{tabular} & \begin{tabular}[c]{@{}c@{}}0.196 $\pm$ 0.05\\ 0.032 $\pm$ 0.01\end{tabular} \\ \hline
					Sensor &\begin{tabular}[c]{@{}c@{}}NMI\\ ARAND\end{tabular} & \begin{tabular}[c]{@{}c@{}}\textbf{0.723 $\pm$ 0.00}\\ \textbf{0.192 $\pm$ 0.00}\end{tabular} & \begin{tabular}[c]{@{}c@{}}0.151 $\pm$ 0.03\\ 0.074 $\pm$ 0.02\end{tabular} & \begin{tabular}[c]{@{}c@{}}0.274  $\pm$ 0.07\\ 0.034 $\pm$ 0.01\end{tabular} & \begin{tabular}[c]{@{}c@{}}0.060 $\pm$ 0.01\\ 0.006 $\pm$ 0.00\end{tabular} & \begin{tabular}[c]{@{}c@{}}0.032 $\pm$ 0.00\\ 0.032 $\pm$ 0.00\end{tabular} & \begin{tabular}[c]{@{}c@{}}0.024 $\pm$ 0.00\\  0.006 $\pm$ 0.00\end{tabular} \\ \hline
					Covertype &\begin{tabular}[c]{@{}c@{}}NMI\\ ARAND\end{tabular} & \begin{tabular}[c]{@{}c@{}}\textbf{0.509 $\pm$ 0.03}\\ \textbf{0.433 $\pm$ 0.10}\end{tabular} & \begin{tabular}[c]{@{}c@{}}0.113 $\pm$ 0.03\\ 0.165 $\pm$ 0.02\end{tabular} & \begin{tabular}[c]{@{}c@{}}0.310 $\pm$ 0.06\\ 0.254 $\pm$ 0.08\end{tabular} & \begin{tabular}[c]{@{}c@{}}0.048 $\pm$ 0.001\\ 0.002 $\pm$ 0.003\end{tabular} & \begin{tabular}[c]{@{}c@{}}0.482 $\pm$ 0.12\\ 0.198 $\pm$ 0.06\end{tabular} & \begin{tabular}[c]{@{}c@{}}0.295 $\pm$ 0.07\\ 0.339 $\pm$ 0.11\end{tabular} \\ \hline
					HyperPlan & \begin{tabular}[c]{@{}c@{}}NMI\\ ARAND\end{tabular} &
					\begin{tabular}[c]{@{}c@{}}\textbf{0.168 $\pm$ 0.01}\\ \textbf{0.041 $\pm$ 0.00}\end{tabular} &
					\begin{tabular}[c]{@{}c@{}}0.026 $\pm$ 0.00\\ 0.035 $\pm$ 0.00\end{tabular} & \begin{tabular}[c]{@{}c@{}}0.140 $\pm$ 0.03\\ 0.093 $\pm$ 0.02\end{tabular} & \begin{tabular}[c]{@{}c@{}}0.002 $\pm$ 0.00\\ 0.001 $\pm$ 0.00\end{tabular} & \begin{tabular}[c]{@{}c@{}}0.026 $\pm$ 0.00\\ 0.027 $\pm$ 0.00\end{tabular} & \begin{tabular}[c]{@{}c@{}}0.014 $\pm$ 0.01\\ 0.019 $\pm$ 0.00\end{tabular} \\ \hline
			\end{tabular}}%
			\end{center}
		\caption{Comparing IMOC-Stream with different algorithms on real datasets. The first value is the average of 10 repetitions and the value after $\pm$ is the standard deviation.}\label{imoc-results}
	\end{table}

	\begin{table}[H]
		\centering
		\resizebox{\textwidth}{!}{\begin{tabular}{|c|c|c|c|c|c|c|c|}
				\hline
				Dataset & Metrics & IMOC-Stream & StreamKM++ & DStream & DBStream & DenStream & CluStream \\ \hline
				1CDT & \begin{tabular}[c]{@{}c@{}}NMI\\ ARAND\end{tabular} & \begin{tabular}[c]{@{}c@{}}\textbf{0.990 $\pm$ 0.07}\\\textbf{0.970 $\pm$ 0.09}\end{tabular} & \begin{tabular}[c]{@{}c@{}}0.759 $\pm$ 0.03\\ 0.679 $\pm$ 0.02\end{tabular} & \begin{tabular}[c]{@{}c@{}}0.691 $\pm$ 0.10\\ 0.667  $\pm$ 0.14\end{tabular} & \begin{tabular}[c]{@{}c@{}}0.631 $\pm$ 0.28\\ 0.610 $\pm$ 0.30\end{tabular} & \begin{tabular}[c]{@{}c@{}}0.208 $\pm$ 0.05\\ 0.086 $\pm$ 0.05\end{tabular} & \begin{tabular}[c]{@{}c@{}}0.621$\pm$0.06\\ 0.583$\pm$0.09\end{tabular} \\ \hline
				1CSURR & \begin{tabular}[c]{@{}c@{}}NMI\\ ARAND\end{tabular} & \begin{tabular}[c]{@{}c@{}}0.481 $\pm$ 0.00\\ 0.248 $\pm$ 0.01\end{tabular} & \begin{tabular}[c]{@{}c@{}}\textbf{0.534 $\pm$ 0.12}\\ \textbf{0.529 $\pm$ 0.17}\end{tabular} & \begin{tabular}[c]{@{}c@{}}0.136$\pm$0.17\\ 0.041$\pm$0.19\end{tabular} & \begin{tabular}[c]{@{}c@{}}0.031$\pm$0.02\\ 0.02$\pm$0.07\end{tabular} & \begin{tabular}[c]{@{}c@{}}0.150$\pm$0.05\\ 0.017$\pm$0.07\end{tabular} & \begin{tabular}[c]{@{}c@{}}0.409$\pm$0.1\\ 0.384$\pm$0.11\end{tabular} \\ \hline
				4CR & \begin{tabular}[c]{@{}c@{}}NMI\\ ARAND\end{tabular} & \begin{tabular}[c]{@{}c@{}}\textbf{0.957 $\pm$ 0.00}\\ \textbf{ 0.954 $\pm$0.00}\end{tabular} & \begin{tabular}[c]{@{}c@{}}0.705$\pm$0.01\\ 0.497 $\pm$ 0.02\end{tabular} & \begin{tabular}[c]{@{}c@{}}0.804 $\pm$ 0.02\\ 0.793 $\pm$ 0.03\end{tabular} & \begin{tabular}[c]{@{}c@{}}0.868 $\pm$ 0.03\\ 0.881 $\pm$ 0.02\end{tabular} & \begin{tabular}[c]{@{}c@{}}0.183 $\pm$ 0.03\\ 0.006 $\pm$ 0.00\end{tabular} & \begin{tabular}[c]{@{}c@{}}0.502 $\pm$ 0.03\\ 0.408 $\pm$ 0.02\end{tabular} \\ \hline
				GEARS\_2C\_2D & \begin{tabular}[c]{@{}c@{}}NMI\\ ARAND\end{tabular} & \begin{tabular}[c]{@{}c@{}}\textbf{0.654 $\pm$ 0.02}\\ \textbf{0.643 $\pm$ 0.01}\end{tabular} & \begin{tabular}[c]{@{}c@{}}0.543$\pm$0.03\\ 0.449$\pm$0.03\end{tabular} & \begin{tabular}[c]{@{}c@{}}0.160$\pm$0.12\\ 0.154$\pm$0.17\end{tabular} & \begin{tabular}[c]{@{}c@{}}0.001$\pm$0.00\\ 0.0001$\pm$0.00\end{tabular} & \begin{tabular}[c]{@{}c@{}}0.021$\pm$0.02\\ 0.010$\pm$0.01\end{tabular} & \begin{tabular}[c]{@{}c@{}}0.301$\pm$0.02\\ 0.219$\pm$0.02\end{tabular} \\ \hline
		\end{tabular}}%
		\caption{Comparing IMOC-Stream with different algorithms on synthetic datasets. The first value is the average of 10 repetitions and the value after $\pm$ is the standard deviation.}\label{imoc-results2}
	\end{table}

\subsection{Clustering High Dimensional Data}
	The curse of dimensionality refers to various phenomena that arise when clustering data in high-dimensional spaces. Most of the clustering algorithms suffer from the curse of dimensionality. This is due to many factors like the high number of parameters to set or the algorithm's high complexity. To prove our method's effectiveness on clustering high dimensional datasets (HDD), we tested it on 6 HDD's from \cite{DIMsets}. The dimensions (number of features) of these datasets vary from $32$ to $1024$ while the number of instances is $1024$ and the number of classes equal to $16$. We compared our results with different stream algorithms based on NMI and ARAND measures. The results are reported in Table \ref{imoc-hdd_results}. The results show that our method outperforms all the other methods in terms of NMI and ARAND. These results are because our algorithm does not require parameter settings and uses linear genetic functions to enhance the quality, unlike the other algorithms. The pre-defined parameter and the use of costly processes and algorithms (like DBSCAN for DBStream and DenStream) make the algorithms slower and not robust when dealing with HDD. The genetic operators and the update of the solutions in our method are performed linearly, making these functions not costly in the computation time.    
	
	\begin{table}[H]
	\centering
		\resizebox{\textwidth}{!}{\begin{tabular}{|c|c|c|c|c|c|c|c|}
\hline
Dataset & Metrics                                             & IMOC-Stream                                                              & StreamKM++                                                               & DStream                                                                 & DBStream                                                                 & DenStream                                                               & CluStream                                                                \\ \hline
dim032  & \begin{tabular}[c]{@{}c@{}}NMI\\ ARAND\end{tabular} & \begin{tabular}[c]{@{}c@{}}\textbf{0.600 $\pm$ 0.01} \\ \textbf{0.227 $\pm$ 0.01}\end{tabular} & \begin{tabular}[c]{@{}c@{}}0.500 $\pm$ 0.04\\ 0.199 $\pm$ 0.04\end{tabular}  & \begin{tabular}[c]{@{}c@{}}0.051 $\pm$ 0.03\\ 0.021 $\pm$ 0.01\end{tabular} & \begin{tabular}[c]{@{}c@{}}0.421 $\pm$ 0.06\\ 0.139 $\pm$ 0.02\end{tabular}  & \begin{tabular}[c]{@{}c@{}}0.062 $\pm$ 0.02\\ 0.003 $\pm$ 0.00\end{tabular} & \begin{tabular}[c]{@{}c@{}}0.211 $\pm$ 0.02\\ 0.0215 $\pm$ 0.01\end{tabular} \\ \hline
dim064  & \begin{tabular}[c]{@{}c@{}}NMI\\ ARAND\end{tabular} & \begin{tabular}[c]{@{}c@{}}\textbf{0.675 $\pm$ 0.01}\\ \textbf{0.380 $\pm$ 0.00}\end{tabular}  & \begin{tabular}[c]{@{}c@{}}0.546 $\pm$ 0.00\\ 0.252 $\pm$ 0.01\end{tabular}  & \begin{tabular}[c]{@{}c@{}}0.037 $\pm$ 0.01\\ 0.005 $\pm$ 0.04\end{tabular} & \begin{tabular}[c]{@{}c@{}}0.522 $\pm$ 0.02\\ 0.339 $\pm$ 0.01\end{tabular}  & \begin{tabular}[c]{@{}c@{}}0.104 $\pm$ 0.04\\ 0.017 $\pm$ 0.01\end{tabular} & \begin{tabular}[c]{@{}c@{}}0.184 $\pm$ 0.02\\ 0.012 $\pm$ 0.01\end{tabular}  \\ \hline
dim128  & \begin{tabular}[c]{@{}c@{}}NMI\\ ARAND\end{tabular} & \begin{tabular}[c]{@{}c@{}}\textbf{0.691 $\pm$ 0.01}\\ \textbf{0.418 $\pm$ 0.02}\end{tabular}  & \begin{tabular}[c]{@{}c@{}}0.571 $\pm$ 0.04\\ 0.386 $\pm$  0.12\end{tabular} & \begin{tabular}[c]{@{}c@{}}0.136 $\pm$ 0.01\\ 0.056 $\pm$ 0.02\end{tabular} & \begin{tabular}[c]{@{}c@{}}0.531 $\pm$ 0.02\\ 0.321 $\pm$ 0.01\end{tabular}  & \begin{tabular}[c]{@{}c@{}}0.147 $\pm$ 0.05\\ 0.090 $\pm$ 0.03\end{tabular} & \begin{tabular}[c]{@{}c@{}}0.191 $\pm$ 0.01\\ 0.003 $\pm$ 0.01\end{tabular}  \\ \hline
dim256  & \begin{tabular}[c]{@{}c@{}}NMI\\ ARAND\end{tabular} & \begin{tabular}[c]{@{}c@{}}\textbf{0.777 $\pm$  0.01}\\ \textbf{0.487 $\pm$ 0.00}\end{tabular} & \begin{tabular}[c]{@{}c@{}}0.575 $\pm$ 0.07\\ 0.377 $\pm$ 0.16\end{tabular}  & \begin{tabular}[c]{@{}c@{}}0.078 $\pm$ 0.01\\ 0.055 $\pm$ 0.03\end{tabular} & \begin{tabular}[c]{@{}c@{}}0.391 $\pm$ 0.02 \\ 0.265 $\pm$ 0.01\end{tabular} & \begin{tabular}[c]{@{}c@{}}0.147 $\pm$ 0.03\\ 0.045 $\pm$ 0.01\end{tabular} & \begin{tabular}[c]{@{}c@{}}0.171 $\pm$ 0.00\\ 0.004 $\pm$ 0.02\end{tabular}  \\ \hline
dim512  & \begin{tabular}[c]{@{}c@{}}NMI\\ ARAND\end{tabular} & \begin{tabular}[c]{@{}c@{}}\textbf{0.788 $\pm$ 0.01}\\ \textbf{0.540 $\pm$ 0.00}\end{tabular}  & \begin{tabular}[c]{@{}c@{}}0.606 $\pm$ 0.04\\ 0.538 $\pm$ 0.03\end{tabular}  & \begin{tabular}[c]{@{}c@{}}0.115 $\pm$ 0.01\\ 0.073 $\pm$ 0.02\end{tabular} & \begin{tabular}[c]{@{}c@{}}0.329 $\pm$ 0.10\\ 0.478 $\pm$ 0.12\end{tabular}  & \begin{tabular}[c]{@{}c@{}}0.112 $\pm$ 0.12\\ 0.045 $\pm$ 0.09\end{tabular} & \begin{tabular}[c]{@{}c@{}}0.145 $\pm$ 0.01\\ 0.002 $\pm$ 0.02\end{tabular}  \\ \hline
dim1024 & \begin{tabular}[c]{@{}c@{}}NMI\\ ARAND\end{tabular} & \begin{tabular}[c]{@{}c@{}}\textbf{0.855 $\pm$ 0.00}\\ \textbf{0.717 $\pm$ 0.01}\end{tabular}  & \begin{tabular}[c]{@{}c@{}}0.774 $\pm$ 0.02\\ 0.634 $\pm$ 0.02\end{tabular}  & \begin{tabular}[c]{@{}c@{}}0.112 $\pm$ 0.03\\ 0.020 $\pm$ 0.03\end{tabular} & \begin{tabular}[c]{@{}c@{}}0.305 $\pm$ 0.09\\ 0.414 $\pm$ 0.08\end{tabular}  & \begin{tabular}[c]{@{}c@{}}0.110 $\pm$ 0.05\\ 0.041 $\pm$ 0.07\end{tabular} & \begin{tabular}[c]{@{}c@{}}0.150 $\pm$ 0.00\\ 0.005 $\pm$ 0.01\end{tabular}  \\ \hline
\end{tabular}}
\caption{Comparing IMOC-Stream with different algorithms on HDD datasets. The first value is the average of 10 repetitions and the value after $\pm$ is the standard deviation.}\label{imoc-hdd_results} 
\end{table}
	\subsection{Clustering Evolution}
\begin{figure}[htbp!]
		
		\centerline{\includegraphics[width=0.8\textwidth, height=8cm]{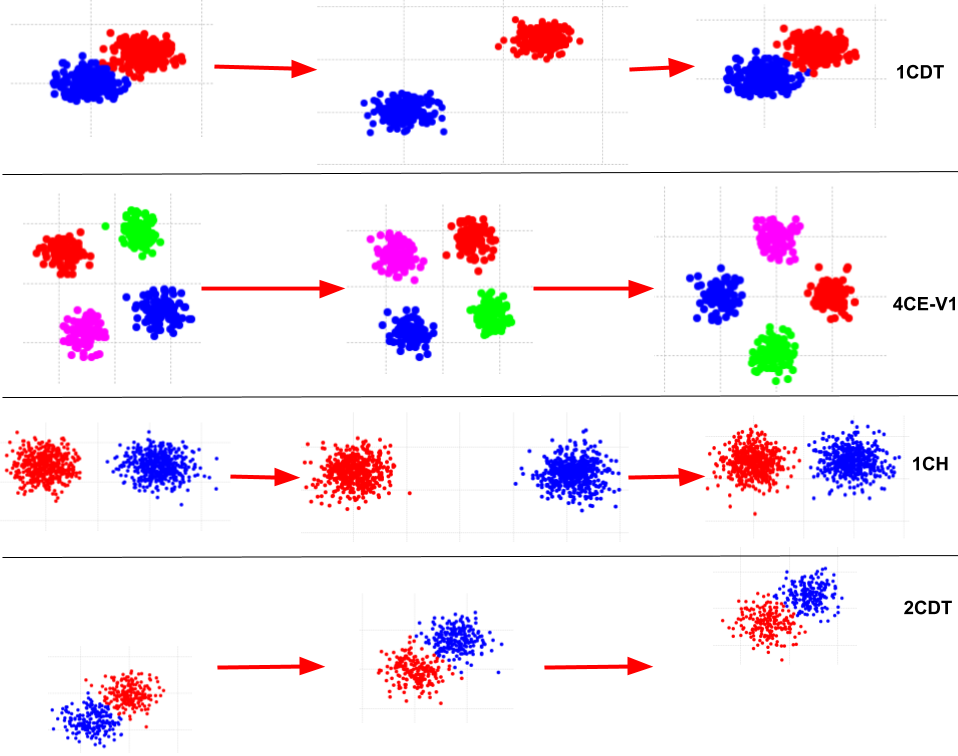}}
		\caption{\label{imoc-visualization} Clustering evolution of 1CDT, 4CE-V1, 1CH, 2CDT datasets. Each color represents a cluster. Each line represents the evolution of a clustering with one dataset.}
	\end{figure}

	\indent Figure  \ref{imoc-visualization} shows an example of the evolution of IMOC-Stream clustering on  {\it 1CDT}, {\it 4CE-V1}, {\it 1CH} and {\it 2CDT} datasets. Each line represents an evolution of clustering for a particular dataset. These figures are generated during the clustering process. We picked three partitionings at random iterations for each dataset. For each time window, the distribution of the incoming data points changes. With its Multi-Objective capability and the fading function's use, IMOC-Stream manages to recognize the structures of the data stream and can separate these structures with the best visualization. It can also detect arbitrarily shaped, compact, and well-separated clusters. We note that the number of clusters does not necessarily stay the same, but the best $K$ is automatically chosen.

		
	\subsection{Arbitrary Shaped Clusters}
	Figure \ref{imoc-arbitrarily} represents the cluster detection for the t4.9k, Compound, and Path-based datasets\footnote{http://cs.joensuu.fi/sipu/datasets/}. We can see from this figure that our method manages to find clusters of arbitrary shapes and provide a good separation of the clusters. The other streaming clustering methods are unable to find clusters of arbitrary shapes (only spherical clusters can be found). The IMOC-Stream method is also able to find noise points due to the use of a density clustering method (DBSCAN).
	
	\begin{figure}[H]
		\centerline{\includegraphics[width=0.8\textwidth]{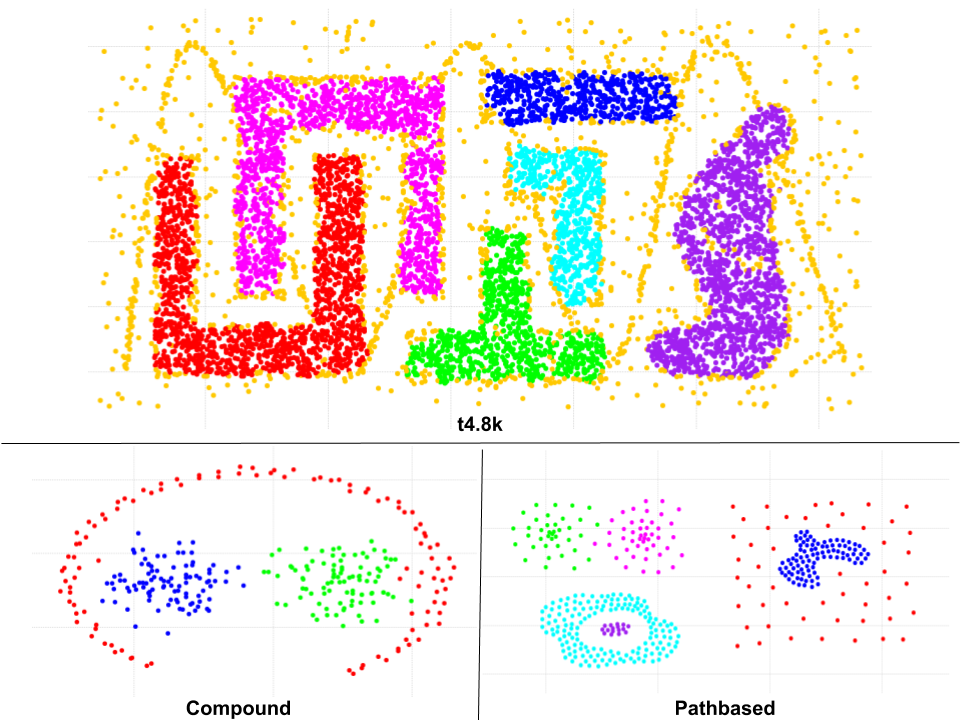}}
		\caption{\label{imoc-arbitrary} Examples of detection of arbitrary shaped clusters by  IMOC algorithm.}
		
	\end{figure}
	\subsection{Time and Memory Complexity}
\begin{figure}[H]
		\centerline{\includegraphics[width=\textwidth]{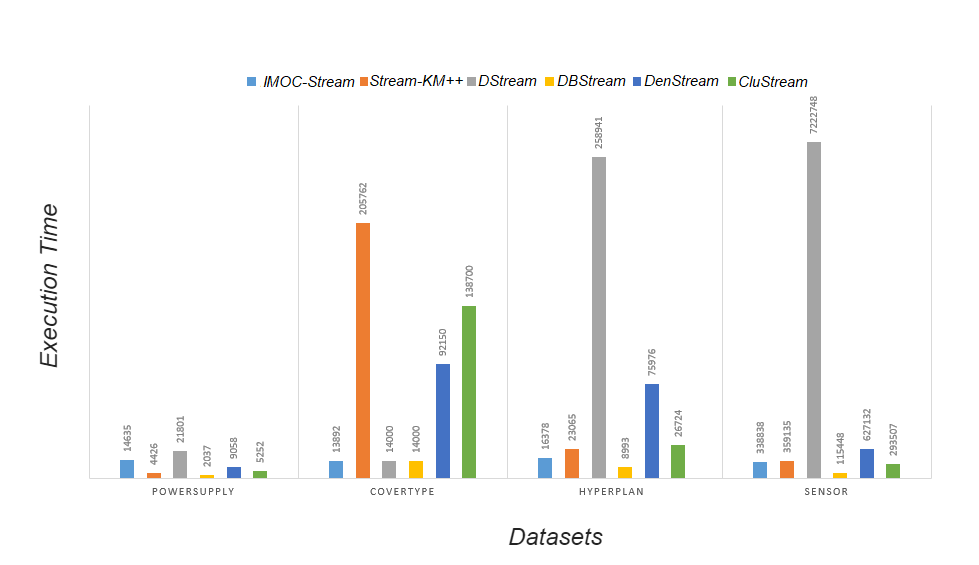}}
		\caption{\label{imoc-time} Execution time in milliseconds of each algorithm for every dataset.}
		
    \end{figure}

\begin{figure}[H]
		\centerline{\includegraphics[width=\textwidth]{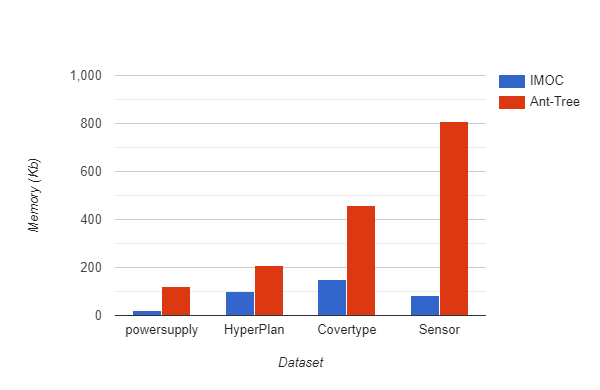}}
		\caption{\label{imoc-memory} Memory allocation in Kilobyte of IMOC-Stream and Ant-tree for every dataset.}
		
	\end{figure}

	In this section, we analyzed our method to prove its improvement over Ant-Tree in terms of memory allocation, and stream algorithms in terms of execution time. Figure \ref{imoc-time} shows the execution times of IMOC-Stream and the other stream clustering algorithms. It can be noticed that the DBStream algorithm has the shortest execution time, but our method is faster than all the different stream clustering algorithms despite its evolutionary nature. In the meantime, IMOC-Stream outperforms all the other algorithms based on the results shown above. These results indicate that IMOC-Stream is non-dominated across all datasets since no different algorithm yields faster computation times. In other words, no other algorithm can produce better results within equal or less time. These results are because our method uses idle times to improve the clustering solutions and use genetic functions with linear complexity instead of using costly functions like the other algorithms.
	Figure \ref{imoc-memory} compares the memory allocation of our method and the Ant-tree algorithm. We observe that IMOC-Stream requires less memory allocation than Ant-tree, on all the datasets. We note that these results are because Ant-tree stores all the data points, making the complexity approximately $n \times d$. While IMOC-Stream stores only the synopsis that is equal to $K \times d$, and when we add the other algorithms' solutions, the memory complexity becomes $\sum_{j=1}^m K_j \times d$, where m is the number of clustering solutions.  

	\section{Conclusion and Future Work}
	\label{imoc-sec:conclusion}
	This paper presents a new method for clustering data stream based on a multi-objective algorithm called IMOC-Stream. Unlike those single-objective clustering techniques that have employed only one objective function, IMOC-Stream employs two objective functions to find clusters of arbitrary shaped clusters and enhance the clustering quality. IMOC-Stream uses a two-phase process: 1) online phase: creating several clustering solutions based on different algorithms and genetic operators 2) offline phase: construction of an optimal partition from the discovered clusters. We applied our method on large stream datasets and compared it to a different stream clustering algorithm. The experiments show the effectiveness of IMOC-Stream for detecting arbitrary shaped, compact, and well-separated clusters with better execution time. Part of our future work should be the proposition of a parallel solution to minimize the execution time. Further research needs to be conducted on incorporating the Ant Colony algorithm since it is suited for parallel algorithms due to its independent agents. More experimentation needs to be conducted using Spark Streaming to test our method on a real stream. We plan to conduct more studies and experiments using different standard optimization algorithms to improve the convergence rate. 
	\section*{Reproductibility}
    To facilitate further experiments and reproducible research, we provide our contributions through an open-source API that contains several clustering algorithms, including: S2G-Stream (local and global version), the 2S-SOM implemented in Spark/Scala and the API documentation at  \href{https://github.com/Clustering4Ever/Clustering4Ever}{Clustering4Ever}\footnote{https://github.com/Clustering4Ever/Clustering4Ever}.
	\section*{Conflict of Intersests}
	The authors declare that they have no known competing for financial interests or personal relationships that could have appeared to influence the work reported in this paper.
	\section*{Acknowledgements}
	A two pages paper from this work has been published as a poster in \cite{attaoui2020moc}. However, this paper is not only a long version of this paper but also introduce an improvement over the previous one.

	%
	%

	\bibliographystyle{apalike}      
	\bibliography{bib3}   
	

\end{document}